\documentclass[runningheads]{llncs}
\usepackage{graphicx}
\usepackage{color}
\usepackage{multirow}
\usepackage{booktabs} 
\usepackage[T1]{fontenc}
\usepackage{algorithm}
\usepackage[noend]{algpseudocode}
\usepackage{caption}
\usepackage{subcaption}

\usepackage{tabularx}

\usepackage{soul} 

\begin{document}
\title{From Robotic Process Automation\\to Intelligent Process Automation\thanks{Authors are in alphabetical order.}\\[1ex]
\textnormal{\em -- Emerging Trends --}}
\titlerunning{RPA to IPA}
%
\author{
Tathagata Chakraborti \and
Vatche Isahagian \and
Rania Khalaf \and
Yasaman Khazaeni \and
Vinod Muthusamy \and
Yara Rizk \and
Merve Unuvar 
}
\authorrunning{T. Chakraborti et al.}
%
\institute{IBM Research AI, Cambridge, MA, USA}
%
\maketitle              

\begin{abstract} 
In this survey, we study how recent advances in machine intelligence are disrupting the world of business processes. Over the last decade, there has been steady progress towards the automation of business processes under the umbrella of ``robotic process automation'' (RPA). However, we are currently at an inflection point in this evolution, as a new paradigm called ``Intelligent Process Automation'' (IPA) emerges, bringing machine learning (ML) and artificial intelligence (AI) technologies to bear in order to improve business process outcomes. The purpose of this paper is to provide a survey of this emerging theme and identify key open research challenges at the intersection of AI and business processes. We hope that this emerging theme will spark engaging conversations at the RPA Forum.

\end{abstract}

\section{Introduction}
\label{intro}

Business processes are an integral part of every industry, such as government, insurance, banking and healthcare. Examples of such processes include automobile insurance claims processing, handling prescription drug orders and patient case management. The business process management (BPM) industry is expected to approach \$16 billion by 2023 \cite{Marketwatch}.
With recent advances in machine learning and artificial intelligence (AI), the automation of steps in a business process -- which came to be known as Robotic Process Automation (RPA) -- is undergoing a radical transformation. 
The industries that are most eager to adopt automation are transportation, manufacturing, packaging and shipping, customer service, finance, and healthcare \cite{McKinsey2017}. 

As noted in ``The Transformation of RPA to IPA: Intelligent Process Automation'' \cite{cmswire}:
{\em The convergence of AI, automation and customer data has now seen the emergence of a new class of tools, known as intelligent process automation (IPA).}
This view is also echoed in market outlook reports from industry leaders, including PwC's recent analysis of rising trends in RPA in the financial sector \cite{ephesoft} and the 2020 AI predictions from IBM research \cite{ibm-sriram} outlining the potential of AI-fueled automation to transform how people work.
Recently AAAI, one of the leading AI conferences, also hosted the first workshop on Intelligent Process Automation \cite{ipa}.
In this survey, we explore this nascent field of inquiry at the intersection of AI and business process automation in greater detail. We begin first with some background on BPM and RPA.

\section{Business Processes} 

\begin{figure}[tbp!]
    \centering
    \includegraphics[width=\linewidth]{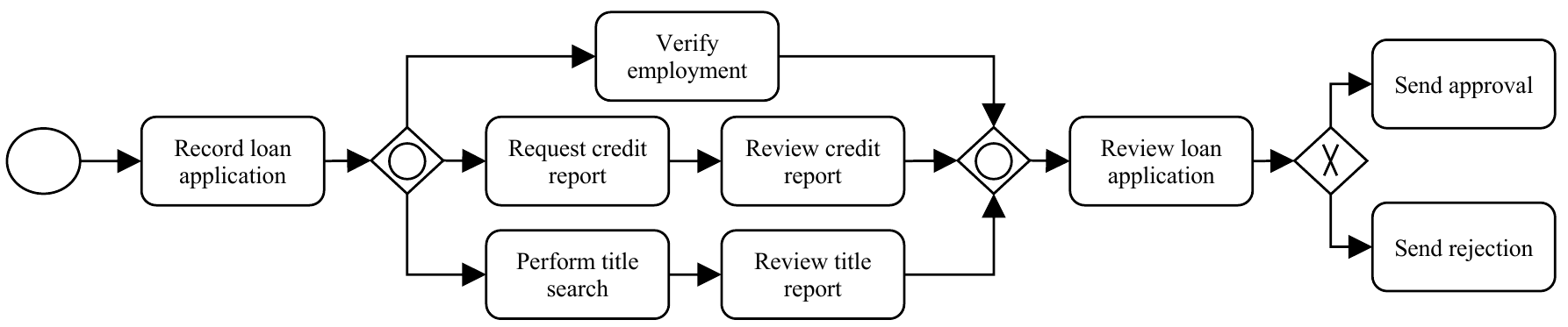}
    \caption{Example of a mortgage loan application process.}
    \label{fig:loan}
\end{figure}

A business process is a collection of connected tasks that once completely executed delivers a service or product to a client or accomplishes an organizational goal within an enterprise \cite{weske2012bpm}. 
A mortgage loan application (shown in Fig. \ref{fig:loan}) is a common example of a business process where the process flow is the set of linked loan application tasks such as collecting client related data (e.g. verifying employment, requesting credit report), performing a title search, receiving the title report and so on. The goal of this process is to approve or reject a loan application once all the required tasks are fully executed. The process is expressed in the business process model and notation (BPMN) graphical notation \cite{grosskopf2009process}. Circles denote events, activities are denoted by rounded-corner rectangles and diamonds depict gateways that allow paths to conditionally merge or diverge.

\subsection{Business Process Management}
Business Process Management (BPM) is a multi-disciplinary field that supports the management of business processes with some combination of modeling, automation, execution, control, measurement and optimization. BPM involves business activity flows (workflows), systems, and people such as employees, customers and partners within and beyond the enterprise boundaries. 

Business processes in reality have a wide scope from the traditional rigid processes (modeled and running under the supervision of a strict workflow management system) to completely ad-hoc unstructured flows driven by humans over e-mail, chat and phone. Traditional BPM systems, at one end of this spectrum, demand a process model that can be completely defined in advance and typically include restrictions such as rigid control flow and context tunneling~\cite{leymannworkflow}. Case management is closer to the other end of this spectrum: a case consists of people, documents, and tasks \cite{weske2012business}. Flexible ordering of task execution is enabled through Event-Condition-Action rules as well as the ability for a user to add new (ad-hoc) tasks. A task itself may be defined as a fully structured workflow, making case management a hybrid model.

\subsection{Business Process Automation}
Businesses seek to support growth while maintaining low costs by automating repetitive and time consuming tasks, especially seeking to eliminate costly and error-prone manual steps. Business Process Automation (BPA) seeks to improve the efficiency of business processes in terms of cost, resources and investment through automating the management of relevant information and data, the time spent by team members, and the execution logic. 

RPA is an emerging technology in BPA that creates software robots that perform tasks previously done by humans. 
RPAs are one form of BPA implementation that is specifically focused on repetitive workflows. The overall goal of RPA is to provide the shortest route to automation by introducing a user interface automation layer rather than interacting deeply with the application code, system or database that are behind those applications.

\subsection{Performance Measures}
Performance measurement in business processes is the first step for analyzing and monitoring the process health and progress for process automation. Identifying the right measures for process performance is extremely important. Performance of a process measures how well the process is doing with respect to chosen indicators. Examples of such performance indicators can be time to execute a task, cost per task in terms of employee head count or number of approved loans \cite{leyer2015}. Numerous authors have proposed a fixed number of category classes for indicators in order to provide a structure. The majority of authors, including \cite{Sarin2011}, proposed a process-oriented view of the indicators which resulted in four groups of performance indicators: quality, time, costs, and flexibility. As \cite{VanderAalst2015} indicates, better processes contribute to meeting the strategic objectives of an organization. Therefore, we specifically call out another association with respect to existing groups of indicators that focuses on the impact of all the indicators towards the business goals. Such indicators can be, for example, analyzing the process performance indicators with respect to profit and revenue of the organization.

In practice, most business process owners focus on productivity measures related to time and cost. The challenge with flexibility and quality measures is that they are difficult to standardize, optimize, implement and generalize. It is common to use indicators related to a process's utilization and assignment of resources, such as repeated tasks or time to execute a task. Resources are usually aligned with the length of the task to avoid bottlenecks in the process. However, shortening the time of task execution or lowering the cost of resources does not necessarily yield better business outcomes. Therefore, it is important to consider methods for measuring process performance that can assess the impact of indicators on business goals and outcomes. 

Adopting the right measures is crucial to the success of BPA. They would then be used to evaluate the performance of RPAs and other automation solutions, allowing business users to assess the effectiveness and return on investment of these solutions. Furthermore, these measures can be used by the RPAs themselves to iteratively improve their performance using machine learning models. 


\subsection{Digital Transformation of Business Processes}
While the use of automation has been gaining traction across many industries, its incorporation into business processes poses several challenges. Automation capabilities such as RPAs can provide transformational benefits, however, it is unclear where their use can provide the highest value. Tools and analytic approaches for identifying high value automation opportunities in a process are still nascent.
As we discussed previously in Section \ref{intro},
recent innovations in machine intelligence 
stand to disrupt this field through the digital
transformation of business processes. 
Throughout the rest of this paper, we will take the 
reader along this journey, starting from the state of the art
in RPAs, through the vision and promise of IPAs, to the 
major challenges to be overcome to reach this goal. 
Finally, we will conclude with a quick overview of
a recently concluded workshop on this topic. 

\section{\mbox{State of the Art: Robotic Process Automation}} 
\label{sec:RPA}

RPA operates on the user interface of software tools and automates mouse and keyboard interactions to remove repetitive, labor intensive tasks. This minimizes human error due to mental lapses resulting from boredom or exhaustion.
RPA's outside-in approach avoided the overhead of changing the internals of legacy software and as a result, its adoption rate has been increasing, leading to its multi-billion dollar valuation. 
Among academic contributions to the field, we distinguish between three approaches to building RPAs.  

The first approach learns to automate tasks by example or demonstration. RPAs either observe humans perform the tasks or process the behavior logs of the software. One example of this approach is if-then-else rule deduction from behavior logs; the form-to-rule approach consisted of identifying the tasks in the logs as humans perform actions on forms and then deduced the rules from the IO data \cite{gao2019automated}. Another example is \cite{le2014flashextract} which provided the input-output examples from which the RPA can extract the underlying rule or program based on the inductive program synthesis paradigm. Miltner et al. \cite{miltner2019fly} detected repetitive edits to text documents by keeping track of a graph of edits and suggested automation rules by adopting a greedy algorithm that finds short explanations of users' edits. All these algorithms rely on humans in the loop. This is one of the more popular approaches to RPAs. However, it does not generalize well to new applications due to the highly specific design of logs and user interfaces. 

The second approach learns tasks from step-by-step natural language text descriptions of the process. Leopold et al. \cite{leopold2018identifying} learned process activities from text documents using supervised machine learning, namely feature extraction using WordNet and support vector machine training (a quadratic optimization approach that finds the optimal separator of activities). Han et al. \cite{han2020automatic} adopted a deep learning approach, long short-term memory recurrent neural networks specifically, to learn the relationship between activities in a business process from text documents describing business processes. This approach also relied on humans in the loop, although in an indirect fashion, since the text documents that describe the processes were written by humans. Since it does not require the existence of an embodied business process (i.e. through a UI), it can be more difficult to learn the rules that should be automated. 

The third approach learns from the task as defined by an environment with its reward function or some input/output examples. Often referred to as RPA 2.0, this approach seeks to eliminate human-dependent training.
It relies on adopting reinforcement learning algorithms on the rewards to RPAs and train them to achieve better performance. This approach is the least mature to date but will lead to generalizable RPAs that approach intelligent automation. 

A critical component to the success of RPAs is identifying the opportunities for automation to add RPAs in the right place and maximize their potential. Bosco et al. \cite{bosco2019discovering} presented a method to analyze user interaction logs in order to discover routines that are fully deterministic and therefore amenable to automate via RPAs. Klingeberg et al. \cite{Klingeberg2018} used process mining to assess the automation opportunities for RPAs with a requirement for processes to be standardized, repeatable and scaleable. Leno et al. \cite{leno2020robotic} proposed a vision for robotic process mining, an approach to achieve end-to-end automation of mining RPA-amenable tasks from logs and generate RPA scripts from these logs to perform the tasks. In practice, these automation opportunities are usually identified manually by subject matter experts comparing potential automation rates. Even though the research in RPA shows promising methods and guidance for assessing automation opportunities, there is still minimal insights on how this could be efficiently automated and implemented in practice at scale. 

\begin{figure}[tb]
    \centering
    \includegraphics[width=\linewidth]{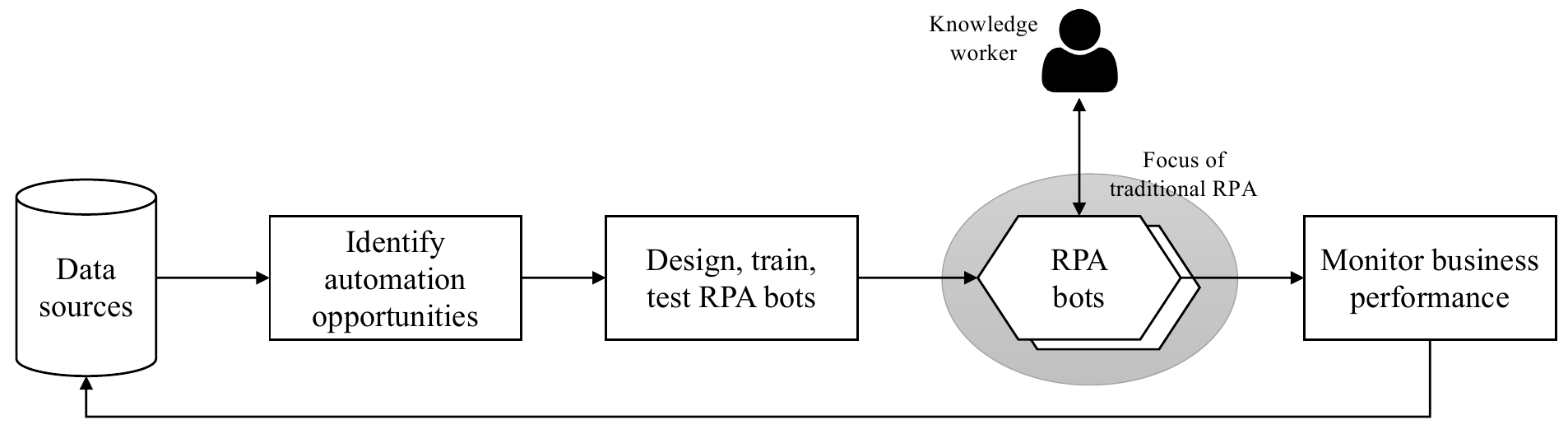}
    \caption{Traditional RPA is focused on building individual bots that automate a repetitive human task. The scope of IPA is broader addressing the coordination between humans and multiple bots, and
    encompassing the entire lifecycle of the process including identifying automation opportunities and continuously retraining the bots based on monitored performance.}
    \label{fig:rpavsipa}
\end{figure}

\section{The Vision: Intelligent Process Automation} \label{sec:vision}

RPA has enabled integration of systems that otherwise would not have been integrated and eased the workload of business process workers automating repetitive and routine tasks (e.g., copying data from one system to another). Beyond automating simple repetitive tasks, IPA achieves more complex automation by using AI to minimize human-dependent training and automating more complex tasks that entail decision making. The IPA vision builds on traditional RPA technologies, while going a step further to automate complex tasks which require decision making, insights and analysis or the composition, coordination, and collaboration of multiple IPA solutions (outside the scope of RPAs as shown in Fig.~\ref{fig:rpavsipa}). While current efforts are a step in the right directions, IPA still falls short of achieving that promise because of the reasons discussed in this section.

\subsection{Automation Opportunities}
Implementing an RPA requires a costly manual analysis of the tasks performed by the users either by observing their behavior, which does not scale when there are hundreds of processes, or through careful analysis of process-related documentation, which can be outdated. Finding opportunities to automate tasks that are more complex than routine repeatable tasks requires the use of structured and unstructured data from process logs. There have been several research efforts to identify candidate tasks for automation \cite{leopold2018identifying,jimenez2019method} from textual descriptions, but they focus on particular business domains (e.g generating utility bills) and are still not implemented at scale. 
Identifying automatable tasks only solves part of the problem. 
The results should also be augmented with a recommendation of possible IPA templates or AI models that are suitable to automate these tasks, as in \cite{leno2020automated}. 

\subsection{High Cost to Build and Maintain}
Unlike RPAs whose overall potential results in a significant increase in turnaround time and cost savings of up to 30\%~\cite{lacity2017new}, there is a higher cost associated with developing IPAs. To build the next generation of IPAs requires data preparation (identifying relevant data, and cleaning and transforming it) and feature engineering (extracting appropriate features), before building and validating the AI capabilities. Similarly, there is a higher cost associated with maintaining IPAs in comparison with RPAs. In addition to the deployed code, the AI capabilities within have a lifecycle of their own.

The AI models must be retrained in response to changes in the business process (control flow drifts) or changes in the data (data drifts). These higher costs require larger return on investment for IPAs to be suitable.
Some ways to mitigate the cost required to build and maintain IPAs include decreasing the effort to develop them, enabling them to be reused for different types of processes, or using them to replace/augment different customer tasks.

\subsection{Low Adoption}
Adopting IPAs comes with an added risk of monetary or reputation loss. For example, data used for training may be manipulated or contain implicit racial, gender, or ideological biases \cite{wolf2017we}. In addition, business users are risk-averse and do not implicitly trust AI models. Mitigating the risk of deploying the AI models requires staged deployment techniques such as canary deployments, bandit services, and A/B testing \cite{muthusamy2018towards}. Increasing business users' trust will require a variety of solutions including maintaining action provenance for audits and providing explanations for any automated IPA decisions.

\subsection{Beyond a Single IPA}
IPA research currently focuses on non-routine tasks. Handling more complex tasks will require the composition of multiple IPAs, as well as the collaboration and coordination of these IPAs. To achieve this, new frameworks need to be developed that enable IPA cooperation. Previous research efforts to use multi-agent systems in BPM
\cite{norman1996designing,coria2014intelligent} 
need to be adjusted and revised for IPAs. Frameworks must now take into account the diversity of automation tasks and domains, and the fact that RPAs can be created by different developers without shared development guidelines. Maintaining compatibility between RPA and business process versions as each co-evolves is also crucial. Finally, a unified interface such as a conversational system may also be required to facilitate the interaction among IPAs and end users \cite{Rizk2020AUnified}. 





\section{Research Opportunities} \label{sec:challenges}





This section highlights research from the BPM and AI literature
to achieve the IPA vision in Section \ref{sec:vision}, 
and outlines opportunities for future research.

\subsection{Business Process Automation}

The BPM literature offers a variety of AI solutions to cluster process traces~\cite{nguyen2016process,nguyen2019summarized} for better process discovery, predict business outcomes~\cite{breuker2016comprehensible,geetika2015}, and provide decision support~\cite{mannhardt2016decision}. Deep learning models, including those in the NLP domain have also been applied~\cite{tax2017predictive,evermann2017predicting}. Recent efforts attempt to discover automatable routines from user interaction logs~\cite{bosco2019discovering}. Unfortunately, due to the reasons mentioned above, very little of these innovations have been applied and adopted by enterprises~\cite{daugherty2018human}, and those adopted are limited to narrow domains such as customer service, enterprise risk, and compliance~\cite{wilson2016companies}. Solutions need to take into account the structure of these highly regulated domains that require paper trails of all transactions and must adhere to privacy and security laws.



\subsection{Composition and Synthesis}

An area of AI that is readily applicable to 
business processes is automated planning, which concerns itself with generating 
sequential courses of actions or plans from its declarative
components and thus provides a powerful framework for 
sequential decision making in a BPM system.
\cite{marrella2017automated} provides an overview of existing work and challenges at the intersection of planning and BPM.
Perhaps the most important (and natural) among them is
the specification and synthesis of business 
processes in the form of planning problems~\cite{r2007integrating}: 
the planner composes workflows
on demand automatically based on the components 
specified by the process author or workflow designer. 

A particular area of interest here is that of the 
composition of automated services for the optimization
of a business process~\cite{dong2004similarity,araghi2012customizing,chakraborti2020d3ba}.
This work is motivated by research on ``web service composition''~\cite{papazoglou2003service}.
We refer to~\cite{rao2004survey} for a comprehensive summary of work in this area, while \cite{srivastava2003web,sohrabi2010customizing} provides an overview of many of the challenges involved.

\subsection{Risk Management}

Another key application of planning to business process
management is in the prediction of how different process
components will evolve over time, thereby anticipating
possible risks. 
Generative model-based approaches such as planning are uniquely 
situated to do this, finding applications in
the robustification and adaptation of processes to failures \cite{jarvis1999exploiting},
validation, verification, and monitoring of processes \cite{de2018aligning}, and so on.
A particular useful tool towards achieving this 
is referred to as top-k and diverse planning \cite{katztop}
where a set of solutions are computed instead of a single 
one thereby allowing one to anticipate likely ways a 
process may evolve.
Such approaches have found many applications\footnote{http://ibm.biz/ai-scenario-planning} in enterprise risk management and scenario planning recently.

\subsection{Chatbots}

Reducing the need for direct human involvement with the business process is one of the main goals of automation.  There is a very strong trend of automating people-driven processes to chatbot interactions throughout the industry~\cite{han2019business}. According to Gartner \footnote{https://www.gartner.com/imagesrv/summits/docs/na/customer-360/C360\_2011\_brochure\_FINAL.pdf}, by 2020 customers will manage 85\% of their relationship with the enterprise without interacting with humans. Conversational interfaces apply not only to customer facing businesses but also to employee services such as help-desk and support bots which have been deployed within almost all enterprises. The focus has also expanded to carrying out a business process with a conversational agent~\cite{lopez2019process}, or automating tasks such as placing orders, paying and following up invoices, repetitive data base queries, external service inquiries and automatic analytics and reporting. 

Chatbots bring ease of access to all these applications in one interactive mode. The natural language interaction helps democratize access to data, automation, and analytics for a broader range of business users enabling faster adaptation by users and greater personalization ~\cite{zumstein2017chatbots}.
Chat interactions also serve as a rich source of data to mine for additional automation candidates, closing the loop on bringing intelligence into process automation.

\subsection{Explainability}

Introducing AI into mission-critical business applications can be a risky endeavor.
The software engineering community has developed formal methods to verify the correctness of programs in critical systems~\cite{formalmethods-2009}, but these techniques are not applicable to learned AI models. Approaches to improve the interpretability and explainability of AI models are a more promising avenue~\cite{xaisurvey-2018}. For example, knowing why a model recommended denying a loan to an applicant is important to ensure adherence to anti-discrimination regulations~\cite{goodman2016eu}. These approaches, however, need to be expanded along at least two dimensions. First, existing interpretability techniques such as perturbation-based methods or interpretable proxy models~\cite{xaisurvey-2018} need to be augmented with domain knowledge of the business process, including the control flow semantics, decision rules, and business objects, thereby leading to more complete and accurate explanations~\cite{aitrust-aaai2020}. Second, the explanations need to be targeted at non-technical subject matter experts. Statistical measures of feature importance or Shapley values are useful to data scientists but do not give actionable insights to a loan officer or process owner. The explanations need to be tailored to the business user, including using the business domain vocabulary and concepts as well as taking into account the context of the user's needs and preexisting knowledge.

\begin{table*}
\centering
\scriptsize
\begin{tabularx}{\textwidth}{p{1.1cm}p{3.7cm}X}
\toprule
Paper & Topics from the Survey & Comments \\[2ex] \midrule
\cite{ferreira2020evaluation} &
Performance Measures, \newline Synthesis & 
This paper attempts to provide a formal framework to facilitate end user programming
of IPAs so that they can be synthesized and evaluated in a principled manner.
\\ \midrule
\cite{agostinelli2020towards} &
RPA --> IPA \newline transformation & 
The authors here echo the message of this survey in terms of the transformation of
RPAs into IPAs, and provide a classification of existing RPA tools towards this end.\\ \midrule
\cite{han2020automatic} &
Process Mining & 
This paper focuses on automated discovery of process components from textual descriptions.
The authors use ordered neurons LSTMs with special process-level language models
to capture process information.\\ \midrule
\cite{maurya2020online} &
Process Automation & 
Authors here focus on automation of the procure to pay process (P2P)
by means of similarity measures learned from recordings of 
a case worker's manual workflow.\\ \midrule
\cite{jenkins2020probabilistic} &
Modalities & 
The authors here explore a stochastic model of spatial demand
in a commercial store in order to optimize produce placement.
Approaches based on deep q-learning techniques
provided promising results.\\ \midrule
\cite{shrestha2020high} & 
Modalities & 
This paper utilizes an R-NET with modified attention to translate instructions in English to navigational plans, providing useful insights on the representation of
graphs with known landmarks and natural language annotations. \\ \midrule
\cite{ayub2020robot} &
Process Mining, \newline Modalities & 
This paper explores how an agent can be taught the rules of a game (process) interactively using a combination of demonstration, active learning, and game theory.\\ \midrule
\cite{chen2020monte} & 
Process Mining, \newline Synthesis & 
Authors here attempt to learn data analysis widgets from SQL query logs and optimize the resultant interface using Monte Carlo tree search methods.\\ \midrule
\cite{Rizk2020AUnified} & 
Chatbots, BPA, \newline Modalities &
As discussed in the survey, this paper explores a multi-agent framework that allows the integration of conversation components in a single interface for the end user.\\ \midrule
\cite{ito2020natural} & 
Modalities, RPA --> IPA \newline transformation & 
This paper explores a natural language interface to IPAs to bring down the expertise level required to manage IPAs using semantic parsing techniques.  \\ \midrule
\cite{chakraborti2020d3ba} &
Chatbots,  Explainability \newline Composition / Synthesis & 
This work also focuses on the end user programming in how
complex business processes with conversational components
can be specified declaratively for automated synthesis
and easy debugging.\\ \midrule
\cite{liinteractive} &
Modalities, \newline Process Mining & 
Authors here again highlight the use of natural language as a means of training IPAs but specifically highlight the effectiveness of a multi-model approach using natural language and GUIs. \\ \midrule
\cite{leno2020automated} &
Process Mining & 
This paper revisits the process mining theme and attempts to learn routines where a user transforms data from one form (spreadsheet or web) to another, by using logs of interactions on a GUI.\\ \midrule
\cite{moiseeva2020multipurpose} &
Chatbots, \newline BPA &
Authors here re-emphasize the usefulness of a conversation interface for business process automation, this time using the assistant to augment unstructured resources with additional training data in order to aid in transfer learning.\\ \bottomrule
\end{tabularx}
\vspace{10pt}
\caption{A summary of the $1^{st}$ International Workshop on
Intelligent Process Automation (IPA) at AAAI 2020 (NYC, Feb 2020).}
\label{ipa}
\end{table*}

\subsection{Modalities}

We posit that business process data should be considered a new modality in machine learning, similar to image, text, audio, or video. 
At the very least, it should be treated as a multi-modal domain \cite{multimodal-2019}. 
A non-exhaustive list of the different types of data embodied in a business process includes the control flow (graph structure), 
the execution of a process trace (sequence of events), 
the metadata associated with an event (multi-dimension set of attributes),
references to unstructured documents (images or text), interactions between participants (both graph and time series representations), and the social networks (also graphs).

Many existing techniques exploit one or more of these data structures to extract insights or build predictive models~\cite{nguyen2016process,nguyen2019summarized,mannhardt2016decision,tax2017predictive,evermann2017predicting}. We believe that a more principled approach to unify these different sub-modalities of business processes will accelerate research in this area. Reifying business process data as a distinct modality opens up a number of research questions for the machine learning community including developing novel techniques for representation learning, explainability, and transfer learning for this new modality.

\section{Closing Remarks: State of the Art in IPA}

So far, we have discussed the promise of IPAs and AI challenges
towards realizing that promise. 
We will now conclude with a brief summary of the recently
concluded (inaugural) international workshop on Intelligent
Process Automation (IPA-20) \cite{ipa} at AAAI 2020.
To the best of our knowledge, this is the first workshop of its
kind at one of the major AI conferences, and it perfectly
reflects the current excitement around business process
automation and artificial intelligence. 
Thus, it is worthwhile to explore the proceedings of the
workshop for the latest areas of interest in this field.

Table \ref{ipa} summarizes the papers presented at IPA-20.
It is particularly interesting to observe recurring themes
in the papers from topics discussed so far in this survey.
Popular topics revolve around process mining and 
automation (particularly from natural language), 
automated synthesis and composition of processes
for end user programming, conversational interfaces to 
business processes, and the need to deal 
with multi-modal inputs.
Multiple keynote speakers also touched on the 
importance of synthesis from examples and natural language understanding
in business process automation.

While these topics covered in the proceedings of IPA-20
largely validate the research agenda laid out in the survey 
so far, it also reveals how much exciting work still needs
to be done for the digital transformation of RPAs to IPAs. 
Most importantly, this transformation cannot be successful
without the effective synergy across the BPM community 
\cite{agostinelli2019research}
and the AI crowd at conferences like AAAI \cite{ipa}, 
which has largely remained separate in spite of the growing
overlap in their interests.
We hope that this survey can act as a springboard for the 
exchange of ideas across the two communities and 
motivates exciting research opportunities going forward, 
combining the power of 
AI and the real-world complexities, challenges, and scale 
of business process automation problems.

%
%
%
\bibliographystyle{splncs04}
\bibliography{references}

\end{document}